\documentclass{article}

\usepackage{arxiv}

\usepackage[utf8]{inputenc} 
\usepackage[T1]{fontenc}    
\usepackage{hyperref}       
\usepackage{url}            
\usepackage{booktabs}       
\usepackage{amsfonts}       
\usepackage{nicefrac}       
\usepackage{microtype}      
\usepackage{lipsum}		
\usepackage{graphicx}
\usepackage{natbib}
\usepackage{doi}

\usepackage{amsmath}
\usepackage{cleveref}
\usepackage{changepage}
\usepackage{anyfontsize}
\usepackage{float}
\usepackage{longtable}
\usepackage{subfig}

\title{Scalable Identification and Prioritization of Requisition-Specific Personal Competencies Using Large Language Models}

\author{
 \textbf{Wanxin Li\textsuperscript{1}\thanks{This work is done during an internship at Amazon.}},
 \textbf{Denver McNeney\textsuperscript{2}},
 \textbf{Nivedita Prabhu\textsuperscript{2}},
 \textbf{Charlene Zhang\textsuperscript{2}}, 
 \textbf{Renee Barr\textsuperscript{2}},\\
 \textbf{Matthew Kitching\textsuperscript{2}},
 \textbf{Khanh Dao Duc\textsuperscript{1,3}},
 \textbf{Anthony S. Boyce\textsuperscript{1}}
\\
 \textsuperscript{1}Department of Computer Science, University of British Columbia, Canada,\\
 \textsuperscript{2}Amazon,\\
 \textsuperscript{3}Department of Mathematics, University of British Columbia, Canada
}

\renewcommand{\shorttitle}{Identification and Prioritization of Requisition-Specific Personal Competencies using LLMs}

\begin{document}
\maketitle

\begin{abstract}
AI-powered recruitment tools are increasingly adopted in personnel selection, yet they struggle to capture the requisition (req)-specific personal competencies  (PCs) that distinguish successful candidates beyond job categories. We propose a large language model (LLM)-based approach to identify and prioritize req-specific PCs from reqs. Our approach integrates dynamic few-shot prompting, reflection-based self-improvement, similarity-based filtering, and multi-stage validation. Applied to a dataset of Program Manager reqs, our approach correctly identifies the highest-priority req-specific PCs with an average accuracy of 0.76, approaching human expert inter-rater reliability, and maintains a low out-of-scope rate of 0.07.
\end{abstract}


\section{Introduction}
\label{sec:intro}
The adoption of AI in personnel selection has accelerated in recent years, with recent surveys indicating that 67\% of the companies now use AI-powered recruitment tools~\cite{chan2025airecruitmnet}. Organizations increasingly rely on technology-assisted systems to screen resumes, predict candidate performance, and support structured interviewing~\cite{koenig2023improving,konig2022machine}. The effectiveness of such systems depends on their ability to identify and prioritize the underlying personal competencies (PCs), which are the knowledge, skills and abilities that drive job success~\cite{maaleki2018arzesh}. Identification ensures comprehensive coverage of job requirements, while prioritization allocates assessment resources on the most important PCs for success in each position.

In organizational hiring, a job category (JC) refers to a broad role type (e.g., Program Manager), while a requisition (req) is a specific job posting for an open position (e.g., a Program Manager supporting a machine learning (ML) team). Standardized PC libraries such as the competency models used at IBM, General Electric, and Google~\cite{vaia2025competency} define PCs at the JC level (e.g., ``Deal with Ambiguity'' for Program Managers). However, individual reqs often require additional PCs with domain expertise or functional capabilities beyond the JC (e.g., ``ML Systems'' for a Program Manager supporting an ML team), which we refer to as req-specific PCs.
These req-specific PCs are critical for enabling more accurate candidate evaluation in assessment systems~\cite{calhr_competencies,headstart_competency_hiring}. 

In this paper, we introduce a scalable Large Language Model (LLM)-based approach to identify and prioritize req-specific PCs. To the best of our knowledge, this is the first work that explicitly distinguishes req-specific PCs from JC-level PCs during a scalable identification and prioritization process~\cite{hilton2010database,workitect_onesizefitsall,valverde2025skill}.  Our approach integrates dynamic few-shot learning~\cite{d2024dynamic}, LLM-as-a-judge~\cite{zheng2023judging} and reflection~\cite{madaan2023self}, similarity-based filtering, and validation against standardized PC libraries.
Throughout development, we implement a human-in-the-loop process where SMEs provide iterative feedback to refine prompts to LLMs. We evaluate our approach on a dataset of Senior Program Manager (PM) and Senior Product Manager-Technical (PMT) reqs from Amazon to assess identification and prioritization accuracy, measure filtering effectiveness, and conduct ablation studies examining the contribution of individual components.

\section{Related work}
Previous work on PC identification and prioritization from reqs can be grouped into the following categories.
\paragraph{Keyword- and ontology-driven approaches} rely heavily on structured taxonomies and predefined vocabularies. Systems such as ONET and ESCO provide foundational frameworks for mapping job requirements to standardized PCs~\cite{levine2013net,handbook2017european}. 
Recent work extends these approaches through normalized skill extraction and semantic similarity techniques to match req text against taxonomies and assign importance scores~\cite{malandri2025skillmo,alonso2025novel}.

\paragraph{Statistical approaches} combine text mining techniques with quantitative methods for skill extraction and prioritization. Several studies apply Term Frequency–Inverse Document Frequency (TF-IDF), text mining, and Pareto analysis to extract skills and rank them by frequency or demand indices~\cite{gavrilescu2025techniques, icsiugiccok2023analysis, darabi2018detecting}. Unsupervised topic modeling methods discover latent skill patterns from job corpora, with approaches using latent semantic analysis and neural topic modeling to extract and rank skills by percentage-based importance~\cite{akkol2023topic, gurcan2025towards, bogdany2023proposed}. 

\paragraph{LLM-based approaches} leverage recent advances in LLMs to significantly improve skill extraction from job postings.
Zhang et al.\ developed domain-adapted transformer models (ESCOXLM-R, JobBERT, JobSpanBERT) for cross-lingual and span-level extraction of hard and soft skills~\cite{zhang2023escoxlm, zhang2022skillspan}. However, these transformer-based methods require substantial labeled training data.
Several studies employ GPT models with various prompting strategies to extract and categorize skills from job postings, then rank them using frequency-based metrics (TF-IDF, occurrence counts) or embedding similarity~\cite{fang2023recruitpro, li2023skillgpt, chumwatana2025bridging, azofeifa2025insights,decorte2023extreme}. Herandi et al.\ fine-tuned general-purpose LLMs (Skill-LLM) to generate skill entities~\cite{herandi2024skill}. 

However, these existing approaches have several limitations. Keyword- and ontology-driven approaches, and statistical approaches, can only extract explicitly mentioned skills. All methods discussed above treat all PCs uniformly, and hence do not distinguish between JC-level PCs and req-specific PCs during identification. For prioritization, frequency-based methods may overweight common but less critical terms, and similarity-based methods are limited to specific reference terms. Additionally, topic modeling approaches often produce difficult-to-interpret clusters.

\section{Methodology}

\begin{figure*}
    \centering
    \includegraphics[width=\linewidth]{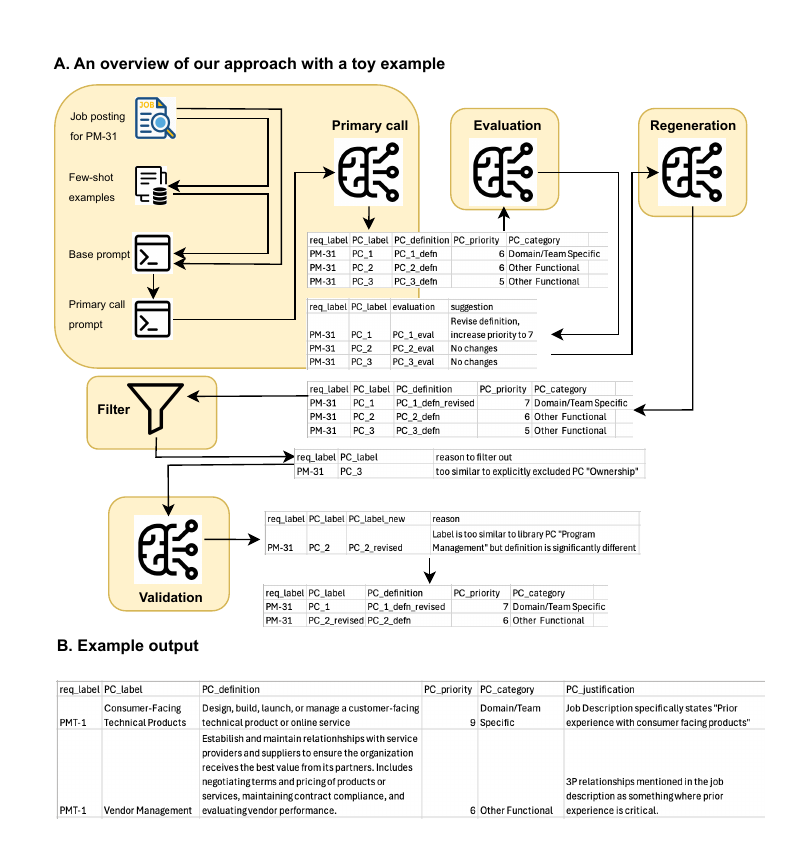}
    \caption{\textbf{(A)} An example output from our approach. In this example, our approach output two PCs for PMT-1: A ``Domain/Team-Specific'' PC ``Consumer-Facing Techinical Products'' with a priority rating of 9 and a ``Other Functional'' PC ``Vendor Management'' with a priority rating of 6. \textbf{(B)} An overview of our approach with a toy example using primary call, evaluation, regeneration, filter and validation components. PC justification is omitted from the output to save space. In this example, we want to identify req-specific PCs for PM-31. In the primary call component, we prepare the primary call prompt from the reqs and the most similar example from the example library. In the evaluation component, an LLM  evaluates the outputs from the primary call and gives a suggestion to revise PC\_2 definition. In the improvement component, an LLM uses the suggestion to improve PC\_2 definition and corrects the priority for PC\_1 using rules. In the filter component, we filter out PC\_3 as it is too similar to ``Ownership'', which is an explicitly defined PC to exclude (out-of-scope). In the validation component, we validate each PC label against the standardized competency library and find PC\_2'label is too similar to the library PC ``Program Management'' but with significantly different definitions. Hence, our label refinement LLM refines PC\_2 to a different label PC\_2\_revised to avoid confusion. }
    \label{fig:components}
\end{figure*}

\Cref{fig:components}.A shows an overview of our approach with the key components with an example. Given a req including job description, our approach outputs a ranked list of req-specific PC labels with their definitions, priority ratings (i.e., 1-10 scale based on importance of mention in reqs), categorical labels (i.e., Domain/Team-Specific for role-specific expertise or Other Functional for general capabilities),\footnote{Details of categorical labels,  priority rating and ranking rules are provided in~\Cref{app:category_priority}.} and justifications citing specific evidence from the req (see~\Cref{fig:components}.B for an example output). 

\subsection{Primary call component}
\label{sec:primary}
The first component uses an LLM with dynamic few-shot learning to generate an initial set of req-specific PCs. We construct an instruction prompt that includes PC label guidelines, PC definition guidelines, out-of-scope PC, categorical label definitions, granularity guidelines, and justification guidelines (see~\Cref{tab:SME-assessed-ratings-defn} and~\Cref{app:rating-dimension-groundtruth} for details). Out-of-scope PCs include JC-level PCs already covered by the standardized competency library and other explicitly excluded PCs based on the specific use case. Then, the component dynamically identifies the most relevant example in the example library by comparing contextual similarities using sentence-transformer embeddings from job descriptions, and selecting the most similar example that exceeds a defined similarity threshold; if no examples meet this threshold, the component defaults to zero-shot learning. The selected example contains a job req and desired req-specific PC output. We add this example to the prompt and send it to an LLM to generate the initial set of PCs.

\subsection{Evaluation component}
In the second component, an LLM takes a prompt that evaluates each PC across the six dimensions mentioned in the previous component. Then, an LLM takes the prompt with the evaluations and generates specific improvement suggestions for each identified issue. 

\subsection{Regeneration component}
In the regeneration component, an LLM takes a prompt with the initial PCs and improvement suggestions from the evaluation component to revise them into a higher-quality set. 

\subsection{Filter component}
The filter component removes redundant and out-of-scope PCs using similarity-based filtering. Redundant PCs are those that are too similar to higher-priority model-generated PCs. 

\subsection{Validation component}
The final component eliminates confusion between newly-generated PC labels and existing PCs in the standardized PC library. The component computes the similarities between model-generated PCs and library PCs, and uses two actions: (1) direct replacement for similar PCs to existing PCs, and (2) label improvement for req-specific PCs with highly similar labels to existing  PCs but highly different definitions. For label improvement, we use a smaller LLM that takes a prompt to suggest improved labels that maintain the essence of the identified PC while aligning with standard PC guidelines.

\subsection{Prompt refinement}
\label{sec:prompt_refinement}
The prompts used in the primary call, evaluation, regeneration and validation components are refined using a human-in-the-loop process (see~\Cref{app:human_in_loop} for an illustration). For each batch used for prompt refinement,\footnote{In in~\Cref{sec:dataset}, this refers to the train and dev set.} SMEs review model-generated outputs and provide SME-assessed ratings using the anchors defined in~\Cref{tab:SME-assessed-ratings-defn} (see~\Cref{app:rating-dimension-groundtruth} for details), and qualitative feedback on model outputs, which were used to update prompts for the next iteration unless reaching target thresholds.

\begin{table}[t]
\begin{tabular}{|p{0.4\linewidth}|p{0.28\linewidth}|p{0.15\linewidth}|}
\hline
\textbf{Evaluation/ \newline Rating names} & \textbf{Evaluation/ \newline Rating anchors} & \textbf{Used in}  \\
\hline
Out-of-scope check  & Binary (0/1) & Both \\
\hline
{Granularity \newline appropriateness} & Binary (0/1) & Both \\
\hline
{Categorization \newline correctness}  & Binary (0/1) & (b) \\
\hline
{Justification \newline quality} & Binary (0/1) & (b) \\
\hline
{Overlapping \newline meanings check} & Binary (0/1) & (b) \\
\hline
Top-1 appropriateness  & 1-3 scale & (a) \\
\hline
\end{tabular}
\caption{Dimensions used in (a) SME-assessed ratings and in the (b) evaluation component.}
\label{tab:SME-assessed-ratings-defn}
\end{table}

\section{Experimental setup}

\subsection{Data collection}
\label{sec:dataset}
We sampled 140 PM and PMT reqs over Amazon reqs in 2024. We split the dataset into three: train (26\%), used for the example library in~\Cref{sec:primary} and prompt refinement in~\Cref{sec:prompt_refinement}; dev (50\%), used for prompt refinement in~\Cref{sec:prompt_refinement}; and test (24\%), used for evaluation. The label sets differ between the train and dev sets. Ground-truth label sets were created for the train and test sets, where SMEs independently analyzed job reqs. For each req, each SME identified up to 5 req-specific PCs and provided the individual-SME label sets for each identified PC. After completing their individual labeling, SMEs had a consensus meeting, identified matches between individual-SME label sets, and produced one final consensus label set for each req, which we regard as SME ground-truth label sets. Details on this process and output can be found in~\Cref{app:rating-dimension-groundtruth}. The label collection process for the dev set is detailed in~\Cref{app:ground_truth_review}.

\subsection{Evaluation metrics}
\label{sec:evaluation-metrics}
We evaluated our approach using two types of metrics: algorithmic-determined metrics that ensure scalable evaluation, and SME-assessed ratings that offer qualitative human judgment.

For algorithmic-determined metrics, we identified PC matches using a similarity-based approach with bipartite matching to maximize matched pairs within each req (see~\Cref{app:matching} for details). Between the model-generated and SME ground-truth PCs, we calculated top-1/2/3 precision,\footnote{We valued top-1 precision because in the initial launch of this project, we will first use the top-1 req-specific PC for candidate assessment. Top-2/3 precisions provide supporting context.} ranking alignment, priority alignment and categorical alignments. The detailed definitions and formulae for metrics can be found in~\Cref{app:evaluation_metrics_details}.


For SME-assessed ratings, SMEs manually identified matches for each req and provided ratings on dimensions including out-of-scope checks, granularity appropriateness, and top-1 precision using the anchors defined in~\Cref{tab:SME-assessed-ratings-defn} (see~\Cref{app:rating-dimension-groundtruth} for details).

\subsection{Target thresholds}
We set our target thresholds for algorithmic-determined metrics using inter-rater reliability (IRR) scores derived from individual-SME label sets. Specifically, for each req, we calculated pairwise agreement across the metrics, and averaged across all rater pairs to establish benchmark thresholds (see~\Cref{tab:IRR}). 
For SME-assessed metrics, we targeted 0.10 for the out-of-scope rate, as it represents acceptable rates for practical considerations.

\begin{table}
\center
\begin{tabular}{lcccc}
\hline
 & top-1  & ranking  & alignment & category \\
  & precision  & alignment & alignment & alignment \\
\hline
PM & 0.88  & 0.68 & 0.88 & 1 \\
PMT & 0.79 & 0.65 & 0.88 & 1 \\
\hline
\end{tabular}
\label{tab:IRR}
\caption{IRR from SME for algorithmic-determined metrics for PM and PMT reqs.}
\end{table}

\subsection{Implementation details}
We used Claude Sonnet 4~\cite{anthropic2025system} in the primary call, evaluation and regeneration components, selected based on prior successful deployment of Claude models for standardized competency library development within Amazon. We used Claude Haiku 3.5~\cite{anthropic2024claude35haiku} in the validation component. Evaluation and regeneration components run for one iteration. Filter and validation components use batch processing with pre-computed embeddings. We computed similarities using sentence-transformer embeddings (109M params)~\cite{reimers2019sentence}.
Through heuristic analyses on the train set comparing similarity scores with SME ground-truth label sets, we used weights for labels (0.3) and definitions (0.7) for PC similarity, and set a similarity threshold of 0.5 for matching. Explicitly defined PCs to exclude are taken from Amazon's Leadership Principles and Core Competencies. Standardized PC library is Amazon's Job Profile Library. Three SMEs with PhD-level expertise in industrial-organizational psychology conducted data labeling and all human evaluations. While the prompts and data cannot be publicly released due to confidentiality agreements, this multi-component approach remains generalizable to other PC frameworks.

\section{Results}
\subsection{Top-1 PC identification correctness}
\label{sec:top-1-results}
We evaluated the ability of our approach to accurately identify the top-1 req-specific PCs on the test set. For algorithmic-determined metrics in~\Cref{tab:algorithmic-determined-metrics}, we obtained an average top-1 precision of 0.77 with a 95\% confidence interval (CI) of [0.70, 0.84] for PM reqs and an average top-1 precision of 0.75 with a 95\% CI of [0.71, 0.79] for PMT reqs. SME-assessed ratings in~\Cref{tab:SME-assessed-ratings} yield similar results, with a top-1 precision of 0.76 for PM reqs and 0.78 for PMT reqs. These scores approach our target thresholds of 0.88 for PM reqs and 0.79 for PMT reqs in~\Cref{tab:IRR}.
\begin{table*}
\center
\begin{tabular}{lccccccc}
\hline
 & {\small Number of} & {\small top-1} & {\small top-2} & {\small top-3}  & {\small ranking} & {\small priority} & category \\
 & {\small reqs} & {\small precision} & {\small precision} & {\small precision}  & {\small alignment} & {\small alignment} & alignment \\
\hline
PM  & 17 & 0.77 & 0.86 & 0.91  & 0.60 & 0.82 & 0.85 \\
 & & [0.70, 0.84] & [0.81, 0.91] & [0.86, 0.95]  & [0.56, 0.65] & [0.81, 0.84] & [0.81, 0.88]  \\
PMT  & 17 & 0.75 & 0.88 & 0.93 &  0.59 & 0.88 & 0.86 \\
 & & [0.71, 0.79] & [0.84, 0.92] & [0.90, 0.96] & [0.56, 0.62] & [0.87, 0.89] & [0.83, 0.89] \\
\hline
\end{tabular}
\caption{Averages and 95\% CI for algorithmic-determined metrics on the test set. Results are averaged over 10 runs.}
\label{tab:algorithmic-determined-metrics}
\end{table*}

\begin{table*}
\center
\begin{tabular}{lccccc}
\hline
 & Number of ratings  & top-1   & overall & top-1  & granularity \\
 & ratings & {\small precision}  &  out-of-scope rate & out-of-scope rate & appropriateness  \\
\hline
{\small PM} & 282 & 0.76  & 0.07 & 0.06 & 0.87  \\
{\small PMT} & 331 & 0.78 & 0.07 & 0.06 & 0.69  \\
\hline
\end{tabular}
\label{tab:SME-assessed-ratings}
\caption{SME-assessed ratings on the test set. Results are evaluated on 1 run.}
\end{table*}





To analyze the reasons for incorrect top-1 model-generated PCs, SMEs identified whether top-1 SME ground-truth PCs exist in model-generated PCs for each req. 
Analysis of incorrect top-1 PCs reveals that 6 of 8 errors occur because the correct PC is absent from model-generated PCs.
In these cases, our approach mostly generates PCs that are either too broad (e.g., ``Technical Systems'' instead of ``ML Systems'') or too specific (e.g., ``AWS Lambda Development'' instead of ``Cloud Computing''). These errors align with granularity appropriateness in~\Cref{tab:SME-assessed-ratings}, which shows that around 13–31\% of generated PCs need granularity adjustment. The remaining two errors are caused by incorrect priority ratings. We further examined the alignment metrics. While priority alignment scores (0.82 with CI of [0.81, 0.84] for PM, 0.88 with CI of [0.87, 0.89] for PMT) are close to IRR targets (0.88), category alignment are behind (0.85 with CI of [0.81, 0.88] for PM and 0.86 with CI of [0.83, 0.89] for PMT), compared to perfect IRR scores of 1.0. This gap indicates that our approach occasionally misclassifies PCs between the ``Domain/Team-Specific'' and ``Other Functional'' categories.

The performance improves substantially when considering top-2 and top-3 matches: top-2 precision reaches 0.86-0.88 and top-3 precision reaches 0.91-0.93 for both reqs. This indicates that when the model misses the exact top-1 PC, it produces near-top-1 PCs for most cases.
\subsection{PC exclusion scalability}
We evaluated the ability of the model to exclude or deprioritize req-specific PCs that are out-of-scope. We classify a model-generated PC to be an out-of-scope defect if it has a priority rating $\geq$ 6/10 and it is labelled by SMEs as ``out-of-scope''. On the test set, we obtained an overall SME-assessed out-of-scope rate of 0.07 for both PM and PMT reqs, satisfying our target rate of 0.10. In addition, the top-1 out-of-scope rate (percentage of reqs where the top model-generated PC is out-of-scope) is 0.06 for both PM and PMT reqs. The results demonstrate that our approach effectively filters out-of-scope PCs through the combination of the evaluation, regeneration and filter components.
\subsection{Ablation study}
\label{sec:ablation-study}
We conducted ablation studies by (1) experimenting with different prompting techniques (zero-shot and static few-shot), (2) removing components and extended reasoning, and (3) using Claude 3.7~\cite{anthropic2025claude37systemcard}. Results are in~\Cref{app:ablation}.

For prompting techniques, zero-shot learning performs similarly to our full approach (0.02-0.04 drop in top-1 precision). For component and extended reasoning removal, extended reasoning has the largest negative effect on top-1 precision, with removal decreasing top-1 precision to 0.69 for PM reqs and 0.66 for PMT reqs. The evaluation and regeneration, and validation components have the second largest negative impact, reducing scores from 0.77 to 0.74 for PM reqs and 0.75 to 0.74 for PMT reqs. While the filter component shows mixed impacts on top-1 precision, it proves important for prioritization metrics. Specifically, the filter component reduces category alignment from 0.85 to 0.81 for PM reqs and from 0.86 to 0.85 for PMT reqs. For model choice,
using Claude 3.7 instead of Claude 4.0 significantly decreases overall performance, particularly for PMT reqs (0.19 drop in top-1 precision). These results indicate that while evaluation and regeneration, validation, extended reasoning, and model choice are most crucial for top-1 precision, the filter component contributes importantly to PC prioritization.

\section{Conclusion and discussion}
In this paper, we developed and evaluated an LLM-based approach to identify and prioritize req-specific PCs from reqs. Our approach works by explicitly instructing the LLM req-specific versus out-of-scope PCs, detecting and removing out-of-scope PCs. Our approach achieves top-1 precision close to SME IRRs. When the approach misses the exact top-1 PC, it typically identifies near-top-1 PCs, achieving top-2 and top-3 precision of 0.86-0.93. 
The approach also demonstrates strong performance in excluding out-of-scope PCs, remaining well below the 0.10 target rate. 
Ablation studies confirm that extended reasoning, evaluation, regeneration, and validation components are important for maintaining high top-1 precision, while the filter component plays an important role in ensuring prioritization quality. When applied at scale, this approach could potentially save 3,500 SME hours annually for the nearly 7,000 yearly reqs in Amazon's US-based workflow.

In the future, we will explore the following directions.
First, there is room for metric improvement; we will refine the priority rating rules and improve the model’s ability to generate PCs at appropriate levels of granularity. 
Second, we will generalize and apply the approach beyond PM and PMT reqs. This includes applying the approach to other JCs, analyzing the distribution of predicted req-specific PCs, and focusing on top-1 precision and out-of-scope rates. Third, we will explore the reusability of req-specific PC labels and definitions across JCs; this involves analyzing the generated req-specific PCs to identify commonly occurring PCs, clustering them into canonical forms to create a library of reusable req-specific PCs.

\paragraph{Ethical considerations} This approach functions as a decision-support tool for humans, rather than a replacement. In the early stage, SMEs should review and validate all model-generated PCs before use in candidate assessment, preserving the essential role of human expertise and judgment in personnel selection. 
\newpage

\bibliographystyle{unsrtnat}
\bibliography{references}  

@incollection{levine2013net,
  title={O* NET: The occupational information network},
  author={Levine, Jonathan D and Oswald, Frederick L},
  booktitle={The Handbook of Work Analysis},
  pages={281--301},
  year={2013},
  publisher={Routledge}
}

@inproceedings{malandri2025skillmo,
  title={SkiLLMo: Normalized ESCO Skill Extraction through Transformer Models},
  author={Malandri, Lorenzo and Mercorio, Fabio and Serino, Antonio},
  booktitle={Proceedings of the 40th ACM/SIGAPP Symposium on Applied Computing},
  pages={1969--1978},
  year={2025}
}

@article{webber2010similarity,
  title={A similarity measure for indefinite rankings},
  author={Webber, William and Moffat, Alistair and Zobel, Justin},
  journal={ACM Transactions on Information Systems (TOIS)},
  volume={28},
  number={4},
  pages={1--38},
  year={2010},
  publisher={ACM New York, NY, USA}
}

@inproceedings{akkol2023topic,
  title={Topic Modeling for Skill Extraction from Job Postings},
  author={Akkol, Ekin and Olucoglu, Muge and Dogan, Onur},
  booktitle={Iberoamerican Knowledge Graphs and Semantic Web Conference},
  pages={277--289},
  year={2023},
  organization={Springer}
}

@article{koenig2023improving,
  title={Improving measurement and prediction in personnel selection through the application of machine learning},
  author={Koenig, Nick and Tonidandel, Scott and Thompson, Isaac and Albritton, Betsy and Koohifar, Farshad and Yankov, Georgi and Speer, Andrew and Hardy III, Jay H and Gibson, Carter and Frost, Chris and others},
  journal={Personnel Psychology},
  volume={76},
  number={4},
  pages={1061--1123},
  year={2023},
  publisher={Wiley Online Library}
}

@incollection{konig2022machine,
  title={Machine learning in personnel selection},
  author={K{\"o}nig, Cornelius J and Langer, Markus},
  booktitle={Handbook of research on artificial intelligence in human resource management},
  pages={149--167},
  year={2022},
  publisher={Edward Elgar Publishing}
}

@misc{chan2025airecruitmnet,
  author = {Chan, Elton},
  title = {Top 100+ {AI} in Recruitment Statistics for 2025},
  year = {2025},
  month = sep,
  day = {15},
  howpublished = {Second Talent},
  url = {https://www.secondtalent.com/resources/ai-in-recruitment-statistics/},
  note = {Accessed: 2026-02-12}
}

@article{zheng2023judging,
  title={Judging llm-as-a-judge with mt-bench and chatbot arena},
  author={Zheng, Lianmin and Chiang, Wei-Lin and Sheng, Ying and Zhuang, Siyuan and Wu, Zhanghao and Zhuang, Yonghao and Lin, Zi and Li, Zhuohan and Li, Dacheng and Xing, Eric and others},
  journal={Advances in neural information processing systems},
  volume={36},
  pages={46595--46623},
  year={2023}
}

@article{madaan2023self,
  title={Self-refine: Iterative refinement with self-feedback},
  author={Madaan, Aman and Tandon, Niket and Gupta, Prakhar and Hallinan, Skyler and Gao, Luyu and Wiegreffe, Sarah and Alon, Uri and Dziri, Nouha and Prabhumoye, Shrimai and Yang, Yiming and others},
  journal={Advances in Neural Information Processing Systems},
  volume={36},
  pages={46534--46594},
  year={2023}
}

@misc{vaia2025competency,
  title        = {Competency Models},
  author       = {{Vaia Learning Platform}},
  year         = {2025},
  howpublished = {\url{https://www.vaia.com/en-us/explanations/business-studies/operational-management/competency-models/}},
  note         = {Accessed: 2025-09-23}
}

@article{reimers2019sentence,
  title={Sentence-bert: Sentence embeddings using siamese bert-networks},
  author={Reimers, Nils and Gurevych, Iryna},
  journal={arXiv preprint arXiv:1908.10084},
  year={2019}
}

@article{anthropic2025system,
  title={System card: Claude opus 4 \& claude sonnet 4},
  author={Anthropic, AI},
  journal={Claude-4 Model Card},
  year={2025}
}

@techreport{anthropic2025claude37systemcard,
  author      = {Anthropic},
  title       = {The Claude 3.7 Sonnet System Card},
  institution = {Anthropic},
  year        = {2025},
  month       = {February},
  url         = {https://www.anthropic.com/claude-3-7-sonnet-system-card}
}

@misc{anthropic2024claude35haiku,
  author = {Anthropic},
  title = {Claude 3.5 Haiku},
  year = {2024},
  url = {https://www.anthropic.com/claude/haiku},
}

@article{alonso2025novel,
  title={A novel approach for job matching and skill recommendation using transformers and the o* net database},
  author={Alonso, Ruben and Dessi, Danilo and Meloni, Antonello and Recupero, Diego Reforgiato},
  journal={Big Data Research},
  volume={39},
  pages={100509},
  year={2025},
  publisher={Elsevier}
}

@article{gavrilescu2025techniques,
  title={Techniques for transversal skill classification and relevant keyword extraction from job advertisements},
  author={Gavrilescu, Marius and Leon, Florin and Minea, Alina-Adriana},
  journal={Information},
  volume={16},
  number={3},
  pages={167},
  year={2025},
  publisher={MDPI}
}

@article{icsiugiccok2023analysis,
  title={Analysis of Skills and Qualifications Required in Data Scientist Job Postings Based on the Pareto Analysis Perspective Using Text Mining},
  author={I{\c{s}}{\i}{\u{g}}{\i}{\c{c}}ok, Erkan and {\c{C}}elik, Sadullah and Y{\i}lmaz, Dilek {\"O}zdemir},
  journal={EKOIST Journal of Econometrics and Statistics},
  number={39},
  pages={10--25},
  year={2023},
  publisher={Istanbul University}
}

@inproceedings{darabi2018detecting,
  title={Detecting current job market skills and requirements through text mining},
  author={Darabi, Houshang and Karim, Fazle Shahnawaz Muhibul and Harford, Samuel Thomas and Douzali, Elnaz and Nelson, Peter C},
  booktitle={2018 ASEE Annual Conference \& Exposition},
  year={2018}
}

@article{gurcan2025towards,
  title={Towards a Sustainable Workforce in Big Data Analytics: Skill Requirements Analysis from Online Job Postings Using Neural Topic Modeling},
  author={Gurcan, Fatih and Soylu, Ahmet and Khan, Akif Quddus},
  journal={Sustainability},
  volume={17},
  number={20},
  pages={9293},
  year={2025},
  publisher={MDPI}
}

@article{bogdany2023proposed,
  title={A proposed methodology for mapping and ranking competencies that HRM graduates need},
  author={Bogdany, Eszter and Cserhati, Gabriella and Raffay-Danyi, Agnes},
  journal={The International Journal of Management Education},
  volume={21},
  number={2},
  pages={100789},
  year={2023},
  publisher={Elsevier}
}

@article{zhang2023escoxlm,
  title={ESCOXLM-R: Multilingual taxonomy-driven pre-training for the job market domain},
  author={Zhang, Mike and Van Der Goot, Rob and Plank, Barbara},
  journal={arXiv preprint arXiv:2305.12092},
  year={2023}
}

@inproceedings{zhang2022skillspan,
  title={SkillSpan: Hard and soft skill extraction from English job postings},
  author={Zhang, Mike and Jensen, Kristian and Sonniks, Sif and Plank, Barbara},
  booktitle={Proceedings of the 2022 Conference of the North American Chapter of the Association for Computational Linguistics: Human Language Technologies},
  pages={4962--4984},
  year={2022}
}

@inproceedings{fang2023recruitpro,
  title={Recruitpro: A pretrained language model with skill-aware prompt learning for intelligent recruitment},
  author={Fang, Chuyu and Qin, Chuan and Zhang, Qi and Yao, Kaichun and Zhang, Jingshuai and Zhu, Hengshu and Zhuang, Fuzhen and Xiong, Hui},
  booktitle={Proceedings of the 29th ACM SIGKDD Conference on Knowledge Discovery and Data Mining},
  pages={3991--4002},
  year={2023}
}

@article{li2023skillgpt,
  title={SkillGPT: a RESTful API service for skill extraction and standardization using a Large Language Model},
  author={Li, Nan and Kang, Bo and De Bie, Tijl},
  journal={arXiv preprint arXiv:2304.11060},
  year={2023}
}

@article{chumwatana2025bridging,
  title={Bridging the IT skill gap with industry demands: An AI-driven text mining approach to job market trends using large language model},
  author={CHUMWATANA, TODSANAI and Hpone, AKK},
  journal={Journal of Theoretical and Applied Information Technology},
  volume={103},
  number={6},
  pages={2270--2282},
  year={2025}
}

@inproceedings{azofeifa2025insights,
  title={Insights from a Dynamic KSA Taxonomy Framework: Top 10 Wanted Knowledge, Skills, and Abilities for the INFOCOMM Sector in Mexico},
  author={Azofeifa, Jose Daniel and Rueda-Castro, Valentina and Gonzalez-Gomez, Luis Jose and Butt, Sabur and Noguez, Julieta and Ceballos, Hector G and Caratozzolo, Patricia},
  booktitle={2025 Institute for the Future of Education Conference (IFE)},
  pages={1--11},
  year={2025},
  organization={IEEE}
}

@article{herandi2024skill,
  title={Skill-llm: Repurposing general-purpose llms for skill extraction},
  author={Herandi, Amirhossein and Li, Yitao and Liu, Zhanlin and Hu, Ximin and Cai, Xiao},
  journal={arXiv preprint arXiv:2410.12052},
  year={2024}
}

@misc{headstart_competency_hiring,
  title = {Introduction to Competency-Based Hiring},
  author = {{HeadStart.gov}},
  year = {2025},
  howpublished = {\url{https://headstart.gov/human-resources/article/introduction-competency-based-hiring}},
  note = {Accessed: 2026-02-11}
}

@misc{calhr_competencies,
  title = {Competencies},
  author = {{California Department of Human Resources}},
  howpublished = {\url{https://www.calhr.ca.gov/about-calhr/divisions-programs/workforce-development-division/competencies/}},
  note = {Accessed: 2026-02-11}
}

@article{hilton2010database,
  title={A database for a changing economy: Review of the Occupational Information Network (O* NET)},
  author={Hilton, Margaret L and Tippins, Nancy T},
  year={2010},
  publisher={National Academies Press}
}

@misc{workitect_onesizefitsall,
  title = {What is the One-Size-Fits-All Competency Model?},
  author = {{Workitect}},
  year = {2019},
  month = {February},
  howpublished = {\url{https://www.workitect.com/competency-models/what-is-the-one-size-fits-all-competency-model/}},
  note = {Accessed: 2026-02-11}
}

@inproceedings{valverde2025skill,
  title={Skill-based employment taxonomy in the global IT industry 5.0},
  author={Valverde-Rebaza, Jorge and Rodrigues de G{\'o}es, Fabiana and Noguez, Julieta and Da Silva, Nathalia C},
  booktitle={Frontiers in Education},
  volume={10},
  pages={1418184},
  year={2025},
  organization={Frontiers Media SA}
}

@article{d2024dynamic,
  title={Dynamic few-shot learning for knowledge graph question answering},
  author={D'Abramo, Jacopo and Zugarini, Andrea and Torroni, Paolo},
  journal={arXiv preprint arXiv:2407.01409},
  year={2024}
}

@book{maaleki2018arzesh,
  title={The ARZESH competency model: Appraisal \& development manager's competency model},
  author={Maaleki, Ali},
  year={2018},
  publisher={LAP Lambert Academic Publishing}
}

@article{kuhn1955hungarian,
  title={The Hungarian method for the assignment problem},
  author={Kuhn, Harold W},
  journal={Naval research logistics quarterly},
  volume={2},
  number={1-2},
  pages={83--97},
  year={1955},
  publisher={Wiley Online Library}
}

@article{decorte2023extreme,
  title={Extreme multi-label skill extraction training using large language models},
  author={Decorte, Jens-Joris and Verlinden, Severine and Van Hautte, Jeroen and Deleu, Johannes and Develder, Chris and Demeester, Thomas},
  journal={arXiv preprint arXiv:2307.10778},
  year={2023}
}

@article{handbook2017european,
  title={European Skills, Competences, Qualifications and Occupations},
  author={Handbook, ESCO},
  journal={EC Directorate E},
  year={2017}
}
\newpage

\appendix
\section{Categorical label definitions, priority rating and ranking rules}
\label{app:category_priority}
\paragraph{Categorical label definitions} The definitions for Domain/Team-Specific PCs and Other Functional PCs are as follows:
\begin{itemize}
    \item Domain/Team-Specific PCs: A PC is a domain/team specific PC if both of the following criteria are met:
    \begin{itemize}
        \item The PC reflects an experience, understanding, or knowledge of a domain area, and
        \item The PC reflects the focal purpose of the team or role. This is indicated by looking at the external title, the department name, or by using the information that describes the team or role in the job description.
    \end{itemize}
    \item Other Functional PCs: A PC is another functional PC if the PC only reflects an experience, understanding, or knowledge of a domain area. It can be new PCs, or existing job profile library PCs that are not listed as PCs for this specific job code.
\end{itemize}
\paragraph{Priority rating rules} Basic qualifications (BQs), preferred qualifications (PQs) and job descriptions (JD) are three common sections in reqs. BQs are the minimum requirements a candidate must meet to be considered for the position, such as required education or years of experience. PQs are nice-to-have skills or experiences that make candidates more competitive but are not required for the role. JD is the comprehensive overview of the position that outlines key responsibilities, day-to-day duties, and what success looks like in the role. We have the following rules for priority ratings:
\begin{itemize}
    \item If a PC only appears in PQs, its maximum rating is 4;
    \item If a PC only appears in BQs, its maximum rating is 6;
    \item If a PC appears in both the BQs and JD, its minimum rating is 6;
    \item If a PC appears in the BQs, PQs, and JD, its minimum rating is 7;
    \item If a PC appears in the BQs, PQs, and multiple times in the JD, its minimum rating is 8;
    \item If a PC is a Domain/Team-Specific PC, its minimum rating is 7, even if it's only mentioned in JD.
\end{itemize}
\paragraph{Ranking rules} Give a list of PCs, we rank them according to the following rules:
\begin{itemize}
    \item Domain/Team-Specific PCs precede Other Functional PCs;
    \item Within each category (Domain/Team-Specific and Other Functional), PCs are ordered by their priority ratings (highest to lowest).
\end{itemize}
\newpage

\section{Workflow: Human-in-the-Loop}
\label{app:human_in_loop}
\begin{figure*}[h]
    \centering
    \includegraphics[width=\linewidth]{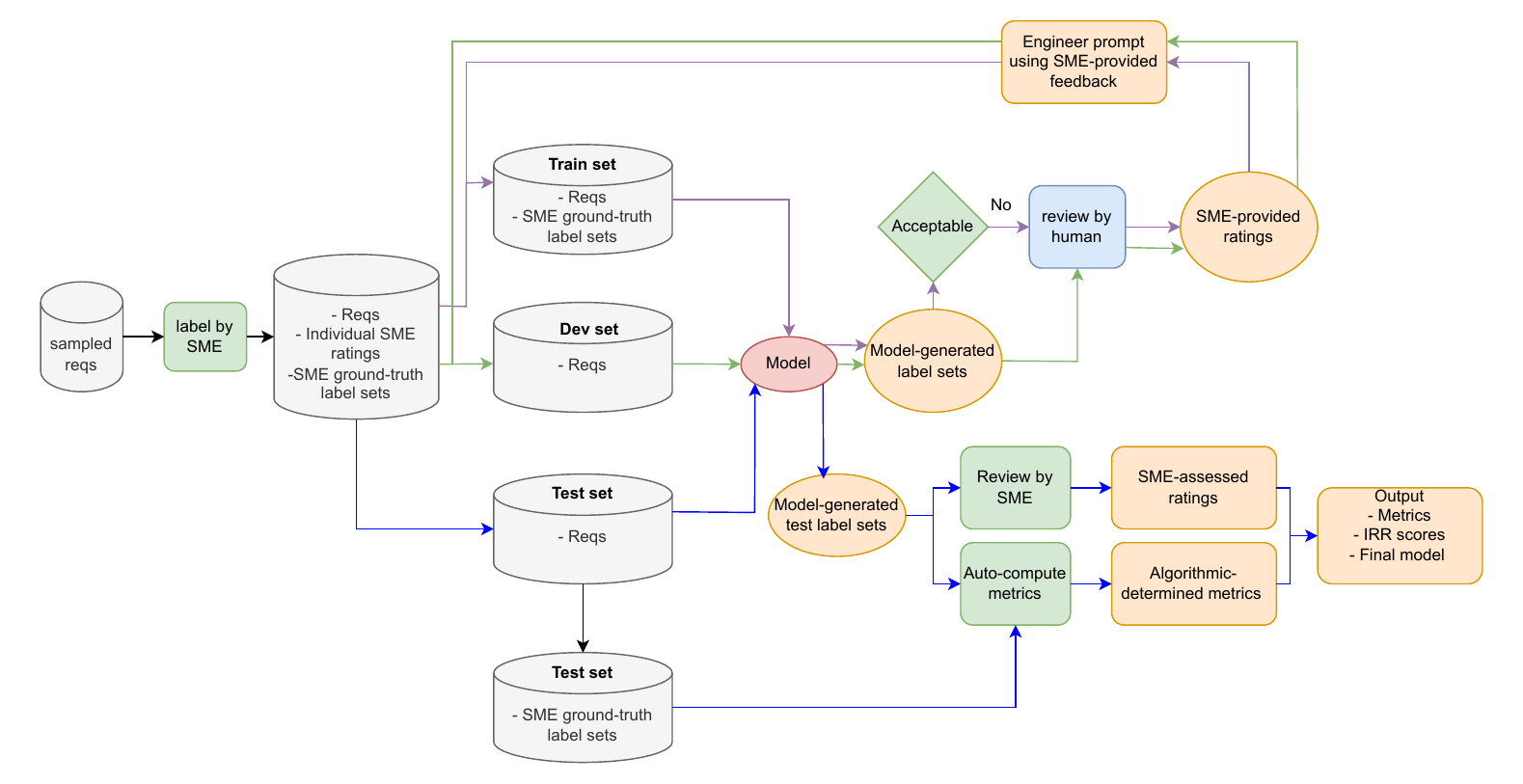}
    \caption{Workflow of req-specific PC identifier. Distinct shapes and colors represent different components and flows. Actions are depicted as blue rectangles, data storage is shown as grey cylinders, model-generated label and evaluation metrics are represented by orange circles, the model itself is indicated by a red circle, and decision points are marked by green diamonds. The pipeline execution follows a sequential order: beginning with steps on the train set (purple lines), followed by steps on the dev set (green lines), and concluding with steps on the test set (blue lines). }
    \label{fig:human-in-the-loop}
\end{figure*}
\newpage

\section{Rating dimension on the SME ground-truth label set collection process}
\label{app:rating-dimension-groundtruth}
To establish ground truth, each SME identified up to 5 req-specific PCs for each req.
We categorized each PC as either domain/team-specific (experience, understanding, or knowledge of a domain area directly related to the team's core purpose) or other functional (any functional competency not considered domain/team-specific). We then prioritized competencies based on their importance to the role, placing domain/team-specific competencies above other functional ones. We then held a consensus meeting to align on a final list of competencies, with associated importance, for each req.

\begin{longtable}{p{0.5\linewidth}p{0.4\linewidth}p{0.1\linewidth}}
\hline
\textbf{Evaluation/Rating Name \& Question} & \textbf{Evaluation/Rating Anchors} & \textbf{Level} \\
\hline
Out-of-scope check - Is this an out-of-scope PC? & 1 - No\newline 0 - Yes\\
\hline
Granularity appropriateness - Does this PC have the just-right level of granularity & 1 - Yes - Just right \newline
0 - No - Too broad \newline
0 - No - Too granular & PC \\
\hline
Categorization correctness - Is the PC categorized correctly as a Domain/Team-Specific or Other Functional? & 1 - Yes\newline 0 - No & PC \\
\hline
Justification quality - Is the PC supported by explicit evidence from the req? &
1 - Yes\newline 0 - No & PC \\
\hline
Overlapping meanings check - Do any PCs have overlapping or redundant meanings? & 1 - No overlap \newline 0 - Redundant/overlapping & Req \\
\hline
Top-1 precision - To what extend does the top-1 model-generated PC match with the top-1 SME ground-truth PC? & 1 - Not appropriate; should not be top-1\newline 2 - Acceptable but not the best\newline 3 - Perfect as top-1 & Req \\
\hline
\caption{Dimensions used in SME-assessed ratings and in the evaluation component, with anchors and level of application (PC vs. req).\label{tab:SME-assessed-details}} \\
\end{longtable}
\newpage

\section{Labels for the dev set}
\label{app:ground_truth_review}

Review label sets are used for the dev set for the dev set where SMEs provided the direct ratings and feedback on model-generated output to guide prompt refinement.
\newpage

\section{Matching process for calculating algorithmic-determined metrics }
\label{app:matching}
For each req, the PC matching process for the algorithmic-determined metrics uses a similarity-optimizing bipartite matching approach. We began the process by calculating similarity between all possible PC pairs, keeping only those exceeding a target threshold. For these valid pairs, the process constructs a cost matrix based on the similarity values. We then applied the Hungarian algorithm~\cite{kuhn1955hungarian} to this matrix to find the assignment that optimizes the total similarity across all matches. The algorithm enforces the constraint that each model-generated PC and each SME-provided PC can be matched at most once, ensuring a one-to-one mapping between the two sets of PCs. 
\newpage


\section{Evaluation metric details}
\label{app:evaluation_metrics_details}
The algorithmic-determined metrics include:
\begin{itemize}
    \item \textit{Top-1 precision}: Percentage of reqs where the model-generated top-1 PC matches the SME ground-truth top-1 PC, considering only top-1 PCs that are Domain/Team-Specific and have priority rating at or above 6/10.
    \item \textit{Top-(2/3) precision}: Percentage of reqs where the model-generated top-1 PC matches one of the SME ground-truth top-(2/3) PCs, considering only top-1 PCs that are Domain/Team-Specific and have priority rating at or above 6/10.
    \item \textit{Ranking alignment (RA)}: Agreement of ranked PC lists where the ranking rules depend on both categorical labels and priority ratings (see~\Cref{app:category_priority} for details) using average rank-biased overlap scores~\cite{webber2010similarity}. Mathematically,
    \begin{equation*}  
    \begin{aligned}
    RA = (1-p)\sum_{k=1}^{\mathop{max}(|M|, |S|)} p^{k-1}\frac{|M_k\cap S_k|}{k},
    \end{aligned}
    \end{equation*}
    where $p=0.9$ is the persistence parameter, $k$ is the depth in ranked PC lists, $M_k$ and $S_k$ are the sets of PCs in the top-k positions from ranked model-generated PC list $M$ and SME ground-truth PC list $S$.
    \item \textit{Priority alignment (PA)}: Agreement of priority ratings and categorical labels between matched pairs between model-generated and SME ground-truth PCs using normalized mean absolute errors. Suppose we have $N$ matched pairs between SME ground-truth and model-generated PCs, and each pair is denoted as $(S^i, M^i)$. Let $p(\cdot)$ denote the priority of a PC. Mathematically,
    \begin{align*}
    PA = 1-\frac{1}{N}\sum_{i=1}^N \frac{|p(M^i)-p(S^i)|}{d}, 
    \end{align*}
    where $d = 10$ is the priority rating scale range. 
    \item \textit{Category alignment (CA)}: Agreement of categorical labels between matched pairs between model-generated and SME ground-truth PCs. Let $c(\cdot)$ be 1 if the categorical label is ``Domain/Team-Specific'' and 0 if ``Other Functional.'' Mathematically, 
    \begin{align*}
    CA = 1-\frac{1}{N}\sum_{i=1}^N |c(M^i)-c(S^i)|.
    \end{align*}
\end{itemize}
\newpage

\section{Ablation study results}
\label{app:ablation}

\begin{table}[h]
\small
\centering
\begin{tabular}{lcccc}
\hline
 & top-1 & ranking & priority  & category\\
  & precision &   alignment & alignment & alignment \\
\hline
Full approach with & 0.77  & 0.60 & 0.82 & 0.85 \\
dynamic few shot (baseline) & [0.70, 0.84] & [0.56, 0.65] & [0.81, 0.84] & [0.81, 0.88] \\
\hline
Zero shot & 0.75 & 0.60 & 0.82 & 0.83 \\
 & [0.70, 0.81]  & [0.57, 0.62] & [0.81, 0.84] & [0.79, 0.87] \\
\hline
Static few shot & 0.57  & 0.55 & 0.83 & 0.80 \\
 & [0.52, 0.64]  & [0.51, 0.58] & [0.82, 0.85] & [0.78, 0.82] \\
\hline
\end{tabular}
\caption{Averages and 95\% CI for algorithmic-determined metrics on the test set for ablation study on experimenting with different prompting techniques for PM reqs.}
\label{tab:PM-ablation-prompt}
\end{table}


\begin{table}[h]
\small
\centering
\begin{tabular}{lcccc}
\hline
 & top-1  & ranking & priority  & category\\
  & precision &    alignment & alignment & alignment \\
  \hline
Full approach with & 0.75  & 0.59 & 0.88 & 0.86 \\
dynamic few shot (baseline) & [0.71, 0.79]  & [0.56, 0.62] & [0.87, 0.89] & [0.83, 0.89] \\
\hline
Zero shot & 0.71  & 0.57 & 0.87 & 0.82 \\
 & [0.66, 0.75] & [0.55, 0.59] & [0.86, 0.88] & [0.79, 0.86] \\
\hline
Static few shot & 0.75  & 0.60 & 0.89 & 0.85 \\
 & [0.70, 0.79]  & [0.57, 0.64] & [0.88, 0.90] & [0.82, 0.88]\\
\hline
\end{tabular}
\label{tab:PMT-ablation-prompt}
\caption{Averages and 95\% CI for algorithmic-determined metrics on the test set for ablation study on experimenting with different prompting techniques for PMT reqs.}
\end{table}



\begin{table}[h]
\small
\centering
\begin{tabular}{lcccc}
\hline

 & top-1  & ranking & priority  & category\\
  & precision &    alignment & alignment & alignment \\
  \hline
Full approach (baseline) & 0.77 & 0.60 & 0.82 & 0.85 \\
 & [0.70, 0.84]  & [0.56, 0.65] & [0.81, 0.84] & [0.81, 0.88]\\
\hline
Without extended reasoning & 0.69  & 0.55 & 0.83 & 0.85 \\
 & [0.65, 0.74]& [0.53, 0.56] & [0.82, 0.84] & [0.82, 0.87] \\
\hline
Without evaluation and regeneration & 0.74  & 0.55 & 0.84 & 0.87 \\
 & [0.68, 0.80]  & [0.53, 0.58] & [0.83, 0.85] & [0.85, 0.89] \\
\hline
Without filter & 0.72  & 0.56 & 0.82 & 0.81 \\
 & [0.68, 0.77]  & [0.54, 0.58] & [0.81, 0.83] & [0.77, 0.84] \\
\hline
Without validation & 0.74  & 0.59 & 0.83 & 0.83 \\
 & [0.70, 0.78]  & [0.56, 0.62] & [0.83, 0.84] & [0.78, 0.88] \\
\hline
\end{tabular}
\label{tab:PM-ablation-component}
\caption{Averages and 95\% CI for algorithmic-determined metrics on the test set for ablation study on experimenting with removing different functionalities and components from the full approach for PM reqs.}
\end{table}
\newpage

  

\begin{table}[h]
\small
\centering
\begin{tabular}{lcccc}
\hline
& top-1  & ranking & priority  & category\\
  & precision &    alignment & alignment & alignment \\
  \hline
  
Full approach (baseline) & 0.75 & 0.59 & 0.88 & 0.86 \\
 & [0.71, 0.79] & [0.56, 0.62] & [0.87, 0.89] & [0.83, 0.89] \\
\hline
Without extended reasoning & 0.66  & 0.56 & 0.89 & 0.85 \\
 & [0.61, 0.72] & [0.54, 0.58] & [0.88, 0.90] & [0.82, 0.88] \\
\hline
Without evaluation and regeneration & 0.74  & 0.56 & 0.88 & 0,85 \\
 & [0.67, 0.80]  & [0.54, 0.57] & [0.87, 0.89] & [0.84, 0.86]\\
\hline
Without filter & 0.76  & 0.59 & 0.88 & 0.85 \\
 & [0.72, 0.80]  & [0.56, 0.62] & [0.87, 0.88] & [0.82, 0.89] \\
\hline
Without validation & 0.74 & 0.61 & 0.87 & 0.83 \\
 & [0.68, 0.79]  & [0.58, 0.63] & [0.86, 0.88] & [0.79, 0.87] \\
\hline
\end{tabular}
\caption{Averages and 95\% CI for algorithmic-determined metrics on the test set for ablation study on experimenting with removing different functionalities and components from the full approach for PMT reqs.}
\label{tab:PMT-ablation-component}
\end{table}


\begin{table}[h]
\small
\centering
\begin{tabular}{lcccc}
\hline
& top-1   & ranking & priority  & category\\
  & precision   &  alignment & alignment & alignment \\
  \hline
Full approach with & 0.77  & 0.60 & 0.82 & 0.85 \\
Claude 4.0 (baseline) & [0.70, 0.84]  & [0.56, 0.65] & [0.81, 0.84] & [0.81, 0.88] \\
\hline
Full approach with & 0.71  & 0.53 & 0.84 & 0.85 \\
Claude 3.7 & [0.64, 0.78]  & [0.51, 0.54] & [0.83, 0.85] & [0.81, 0.89]\\
\hline
\end{tabular}
\caption{Averages and 95\% CI for algorithmic-determined metrics on the test set for ablation study on experimenting with Claude 3.7 from the full approach for PM reqs.}
\label{tab:PM-ablation-model}
\end{table}


\begin{table}[h]
\small
\centering
\begin{tabular}{lcccc}
\hline
 & top-1 & ranking & priority  & category\\
  & precision &  alignment & alignment & alignment \\
  \hline
Full approach with & 0.75  & 0.59 & 0.88 & 0.86 \\
Claude 4.0 (baseline) & [0.71, 0.79] & [0.56, 0.62] & [0.87, 0.89] & [0.83, 0.89] \\
\hline
Full approach with & 0.56 & 0.44 & 0.88 & 0.85 \\
Claude 3.7 & [0.51, 0.62] &  [0.41, 0.47] & [0.87, 0.89] & [0.82, 0.87] \\
\hline
\end{tabular}
\label{tab:PMT-ablation-model}
\caption{Averages and 95\% CI for algorithmic-determined metrics on the test set for ablation study on experimenting with Claude 3.7 from the full approach for PMT reqs.}
\end{table}
\newpage

\end{document}